\documentclass[10pt,letterpaper]{article}

\usepackage{cogsci}

\cogscifinalcopy

\usepackage{pslatex}
\usepackage{apacite}
\usepackage{graphicx}
\usepackage{xcolor}
\usepackage{cancel}

\title{Computational Thought Experiments for a 
More Rigorous Philosophy and Science of the Mind}
 
\author{{\bf Iris Oved (irisoved@gmail.com)} \\
  Independent Scholar, The Paradox Lab \\
  San Francisco, CA 94115 USA
  \And {\bf Nikhil Krishnaswamy (nkrishna@colostate.edu)} \\
  Department of Computer Science, Colorado State University \\
  Fort Collins, CO 80523 USA
  \AND {\bf James Pustejovsky (jamesp@brandeis.edu)   } \\
  Department of Computer Science, Brandeis University \\
  Waltham, MA 02453 USA
  \And {\bf Joshua Hartshorne (joshua.hartshorne@hey.com)} \\
  Department of Psychology, Boston College \\
  Chestnut Hill, MA 02467 USA}

\begin{document}

\maketitle

\begin{abstract}
We offer philosophical motivations for a method we call \textit{Virtual World Cognitive Science} (VW CogSci), in which researchers use virtual embodied agents that are embedded in virtual worlds to explore questions in the field of Cognitive Science. We focus on questions about mental and linguistic representation and the ways that such computational modeling can add rigor to philosophical thought experiments, as well as the terminology used in the scientific study of such representations. We find that this method forces researchers to take a god's-eye view when describing dynamical relationships between entities in minds and entities in an environment in a way that eliminates the need for problematic talk of belief and concept \textit{types}, such as \textit{the belief that cats are silly}, and \textit{the concept CAT}, while preserving belief and concept \textit{tokens} in individual cognizers’ minds. We conclude with some further key advantages of VW CogSci for the scientific study of mental and linguistic representation and for Cognitive Science more broadly. 

\textbf{Keywords:} 
philosophy, methods; concepts; virtual worlds; embodiment; grounding; mental representation; AGI, LLMs.
\end{abstract}

\section{Introduction}
\label{sec:intro}

This paper offers philosophical motivations for a method we call \textit{Virtual World Cognitive Science} (VW CogSci), in which researchers use virtual embodied agents that are embedded in virtual worlds to explore questions in the field of Cognitive Science. We offer a general defense of this method and then focus on the study of mental and linguistic representations. 
\newline

First we recall some of the virtues of computational modeling in general, and then consider reasons to treat cognition as \textit{embodied} (to include sensory receptors and motion actuators) and \textit{embedded} (in a world to be sensed, represented, and acted upon). We then show how the method of VW CogSci allows researchers to go beyond the study of \textit{actual} minds and how they operate in \textit{actual} environments, to the study of various \textit{possible} minds and how they might perform in various \textit{possible} environments, with few practical and ethical constraints.

We then turn specifically to the study of mental and linguistic representation for the remainder of the paper. We summarize some of the persistent philosophical puzzles about beliefs and concepts, which many theorists either ignore or go through painful contortions to accommodate. These include puzzles from Saul \citeA{kripke1979puzzle} and Hilary \citeA{putnam1975meaning}. We then argue that similar puzzles arise not only in far-out philosophical scenarios, but for the common, everyday development of concepts and beliefs, and thus must be taken seriously in a study of cognition. We illustrate this by describing ‘computational thought experiments’ in which two young children hear and then see what in fact are some coyotes and wolves, and try to carve up the categories in their environment to best explain their experiences. We show how VW CogSci dissolves the philosophical puzzles by providing a god's-eye view that eliminates the need for problematic talk of belief and concept \textit{types}, such as \textit{the belief that cats are silly}, and \textit{the concept CAT}, while preserving belief and concept \textit{tokens} in individual cognizers’ minds. We also show how the method allows for a more rigorous science of the mind via the simulation of complex dynamical relationships between mental entities and entities in an environment.

\section{Related Work}
Technologies in recent decades have enabled a method we call \textit{Virtual World Cognitive Science} (VW CogSci), in which researchers build virtual agents in virtual worlds to test by simulation hypotheses about how minds and environments might interact. This is a method that has been used in cognitive robotics \cite{brockman2016openai,oudeyer2007intrinsic,lungarella2003developmental, article, cangelosi2015developmental,9962192}, and includes work from our team \cite{BAW,Krishna2022,pustejovsky2019situational,Puste2022,GhafKrishna2022,ghaffari2023grounding}. Let us recall how we got here.

\subsection{Computational Models in Cognitive Science}

Computational models have been used in the study of complex systems across scientific disciplines --Physics, Biology, Chemistry, Geology, Medicine, Engineering, Economics, and, of course, Cognitive Science. They have been especially powerful in the study of nonlinear systems that are difficult to track with intuition or analytical thinking alone \cite{grubb2020physics}. Experiments are done by building different models, adjusting their variables, running the simulations, and observing the outcome. As technologies develop over time, the models provide more precise and accurate predictions and explanations of observed data. 


In the case of Cognitive Science, computational modeling has been at the core from the start, with its founding claim that cognition is a form of information processing, coupled with the emergence of the field of Artificial Intelligence (AI). First this came in the form of Symbolic AI \cite{turing1950computing,mccarthy1955proposal,newell1961gps}, further defended in Philosophy by \citeA{chomksy1959review}, \citeA{putnam1967Predicates}, and \citeA{fodor1975LOT}, which later came into competition with Neural Network models, beginning with \citeA{mcculloch1943logical} and \citeA{Rosenblatt_1957_6098}. This lead to debates about the architecture of the human mind \cite{churchland1981eliminative,fodor1988connectionism}, continuing today to include Bayesian/Causal approaches \cite{pearl2000models,tenenbaum2011grow}.

However, the field of Cognitive Science defines itself as the study of the mind, which applies broadly not only to \textit{actual} human and animal minds, but also potentially to alien minds, plant minds, silicon-based minds, and all other \textit{possible} minds. What kinds of entities can be minds? Can there be ‘pure’ minds, as \citeA{chalmers2023does} claims, with no sensory connection to a world outside of it, or is sensory grounding required for having thoughts? What kinds of entities can a mind without language think about? Can minds like ours learn categories in a world with much fewer, or vastly different, regularities than the world we live in? Would we understand our world differently if our eyes were on our feet instead of our heads? These questions are relevant to a full understanding of the nature of the mental, what various minds can think about and the roles of sensation, action, and the environment. 

\subsection{Embodied and Embedded Cognition}

The idea that the body and environment are crucial to cognition dates back arguably at least to Aristotle, but was also promoted in the last century by \citeA{husserl1929cartesian}, \citeA{Merleau-Ponty1962-MERPOP-9}, and \citeA{heidegger1975basic}, and was recently revived \cite{gibson1966senses, smith1993dynamic, hutchins1995cognition, clark1997dynamical, lakoff1999philosophy, zahavi2005subjectivity, gallagher2006body, ThompsonEmbodiment2010}. Roughly, the idea of \textit{embodied} cognition is that accounts of cognition must include not only what happens inside the skull, but also what happens in an agent's sensory receptors and motor actuators. The idea of \textit{embedded} cognition is that accounts must also include features of the world being sensed, represented, and acted upon. In the field of AI this pair of claims has come to be known as the symbol-grounding problem \cite{harnad1990symbol}, which was raised originally as a challenge to Symbolic AI, but holds for most Neural Network and Bayesian/Causal models as well.

On our more nuanced account, being causally connected with an external world is neither necessary nor sufficient for cognition. What we require is that the agent has sensori-motor representations that it \textit{treats} as having arisen externally and that it tries to explain with a model of that external world. A brain in a vat, while in fact cut off from the world, can have meaningful thoughts as long as it meets this criterion.  This does, however, rule out Large Language Models (LLMs), like OpenAI's GPT and even so-called `Multimodal LLMs', like GPT-4V, LLaVA~\cite{liu2024visual}, and LLaVAR~\cite{zhang2023llavar}.  
Some (e.g., \citeA{chalmers2023does}) argue that these systems are grounded because their linguistic data come from human users and/or their visual data come from cameras. But they don't \textit{treat} their strings of text as utterances with communicative intent or their images as having been caused by anything outside of themselves. Their processing thus isn’t \textit{cognitive} because it isn’t representational; it isn't aimed at being \textit{about} anything (see \citeA{bender-koller-2020-climbing} and \citeA{harnad2024language} for a more complete defense).

One approach to building embodied and embedded cognitive agents is to build physical, hardware sensors and/or motors to take information from the outside world and to act upon it--- i.e., \textit{robots}. While we feel that this is a step in the right direction, the field of Robotics is far from replicating human or animal vision, or any of our other sensory capacities, and it is far from creating life-like locomotion. Moreover, even as implementations of some possible minds, they are embedded only in our actual world. As we’ve been arguing, we need to experiment not only with how various possible minds might interact with the \textit{actual} world; we want to explore how they would interact with the various \textit{possible} worlds in which they might be embedded. 

\subsection{Virtual Agents in Virtual Worlds}

This brings us to virtual robots, also known as \textit{softbots}. Indeed, one way the field of Robotics has bypassed some of its engineering hurdles is by building virtual robots in virtual worlds. Engineers do this primarily to lower costs during the design of their physical robots. Because of this aim, they use virtual worlds that model the physics of our world, at least the aspects of our physics that are relevant to the functioning of their robots. The video game industry has recently made such real-world-like virtual environments available (see \citeA{quteprints209581} for a review).

A growing number of researchers are using virtual embodied agents in virtual worlds for the study of cognition, particularly for modeling cognitive development in human toddlers. This includes the MIMo (Multimodal Infant Model) project being developed by Jochen Triesch's team \cite{9962192}, which is an open-source platform that embeds a multimodal virtual toddler, with binocular vision, a vestibular system, proprioception, and touch perception, in a virtual world that uses the MuJoCo physics engine. A similar open source platform, VoxWorld, is developed and maintained by members of our team \cite{Krishna2022}, using VoxML (Visual Object Concept Modeling Language) to build on top of the Unity game engine a library of natural-kind objects and artifacts, with various shapes, sizes, surface properties, density, and afforded behaviors \cite{pustejovsky-krishnaswamy-2016-voxml,Puste2022}, including {\it habitats} or configurations that condition such affordances ~\cite{pustejovsky2013dynamic}. Because the platform is built on Unity, which visualizes objects from a player's point of view, it can be easily interfaced with a virtual agent that interacts with the objects in that world through its virtual sensors and movement actuators.

The VoxML platform allows researchers to design a range of possible agents to be embodied and embedded in various VoxWorlds. We can experiment with different `innate' sensory and motor abilities, learning algorithms, memory capacities, and innate theories of physics, biology, language, and other minds. Current extant agents include humanoid, simulated robotic, and self-exploring virtual toddler-like agents. These agents use the hybrid `Best of All Worlds (BAW)' architecture being developed by our team to include the most promising elements of Symbolic, Neural-Network, and Embodied AI \cite{Krishna2022, BAW}. The agents explore objects in their world, taking perceptual samples, and learning about objects' or events' intrinsic or extrinsic properties using various types of machine learning~\cite{pustejovsky2017object,krishnaswamy2017monte,krishnaswamy2018evaluation,ghaffari2023grounding}.\footnote{A philosophically curious feature of current versions is that they are dualistic in that the software running the agent’s algorithms is outside of the virtual world the agent is `embedded' in.} We leave open whether such simulated minds constitute \textit{synthetic} minds, or are \textit{mere} simulations, analogous to simulated hurricanes, to be used in theorizing (see \citeA{Searle1980-SEAMBA}).

The Unity-based VoxWorld platform also allows for the creation of a range of virtual worlds to be used in experimentation. VoxWorlds can be built with very different gravity from our world, or animals that are superficially similar to one another but have different dispositions, or even `gruesome' worlds where objects change their perceptible properties at arbitrary times \cite{Goodman1965-GOOFFA}. This flexibility allows researchers to explore questions about how a given type of mind (e.g., one like ours) might interact with worlds that are quite different from ours, what types of worlds they could learn in, understand, and talk about.\footnote{It might be tempting to suppose that LLMs plausibly maintain something like a virtual world in this sense, however their world model would be internal to the agent, akin to a mental model, not an outside world in which the agent is embedded and trying to model. Moreover, it can be empirically demonstrated that they lack coherent world models. For example, \citeA{ghaffari2024exploring} found that while they can correctly describe a ball or a coconut, they are unable to reason about the effect of a round object on the stability of a structure}

\section{Philosophical Puzzles about Representation}

In this section, we recall three well-known philosophical puzzles about mental and linguistic representation. Later we will show how VW CogSci can help dissolve these puzzles by giving a view from the outside of entities in minds, entities in an environment, and relationships between them. \newline

Even before the emergence of Cognitive Science, philosophers have puzzled over the nature, meanings, and acquisition of mental and linguistic representations. In ordinary parlance, we say things like, ‘Abby hid when she heard a coyote because she \textit{believed that coyotes are monsters} and she \textit{desired that she stay safe}’. Several cognitive scientists have urged the elimination of such propositional attitudes (beliefs and desires) from scientific explanations of human behavior as they fail to map easily onto observable features of the brain and may be better understood as dispositions to behave than as explicit representations \cite{churchland1981eliminative,stich1983folk,dennett1989intentional}. Others have defended them as genuine mental entities, complex representations constructed in part from concepts, like COYOTE, MONSTER, and SAFE, as they explain reasoning, the systematicity and compositionality of thought, and symbolic language use \cite{fodor1988connectionism, Quilty-Dunn_Porot_Mandelbaum_2023}. 
    
But even among realists about propositional attitudes and concepts, puzzles persist when it comes to what we should count, e.g., as instances of \textit{the belief that coyotes are monsters} or \textit{the COYOTE concept}. Part of the problem, we suggest, is the assumption that such representations can be identified by their meanings or ‘content’ --- i.e., by what they represent. What makes it the case that a given mental entity is a representation \textit{of} the property/kind \textit{coyote}?
Internalists \cite{Frege1948-FREOSA-2, Rosch1978-ROSPOC, Segal2000-SEGASB, Prinz2002-PRIFTM} hold that what makes a given mental entity a token of the COYOTE concept is its relation to other representations --- of their furriness, four-legged-ness, distinct howl, and beliefs about their being wild, mammals, and so on. Externalists \cite{putnam1975meaning, Kripke1980-KRINAN, Fodor1998-FODCWC, Burge2010-BUROOO}, in contrast, hold that what makes something a token of the COYOTE concept is that it tends to be caused/activated by instances of coyotes in the outside world. Other theorists \cite{Block1998-BLOCRS-2, Chalmers2006-CHATS}, hold hybrid accounts, on which both internal and external factors are relevant. Others, still, hold that such concepts are constructed gradually over the course of development \cite{Carey2009-CARTOO-3, gopnik-wellman-constructivism}, leaving the question of concept identity indeterminate.\footnote{Indeed they leave concept types indeterminate in ways that align quite well with our position here.} Philosophers, mostly using the same old methods of armchair thought experiments that have been used for millennia, have created increasingly convoluted variations on these accounts \cite{Prinz2002-PRIFTM, Fodor1998-FODCWC, Laurence1999-LAUCAC-3}, while others  (e.g., \citeA{Machery2009-MACDWC}) take the accounts to be so convoluted that we should return to the complete elimination of beliefs and concepts from our science of the mind. 

Next, we describe three philosophical puzzles that persist despite attempts to characterize concept and belief types. The puzzles are often dismissed as edge cases, but we will see later that similar cases are central to accounts of mental representation, particularly of their dynamical interactions with environments during the course of development. 

    \vspace*{-1mm}
\subsection{Kripke’s Case of Pierre in Londres} 
In Saul \citeauthor{kripke1979puzzle}'s \citeyear{kripke1979puzzle} \textit{A Puzzle about Belief}, he describes the example of Pierre, who lives in France and hears about a beautiful city named ‘Londres’, which is the French name for London. Pierre tends to make statements like, ‘Londres est jolie’, which, as Kripke notes, we readily translate and ascribe to Pierre \textit{the belief that London is pretty}. But as the thought experiment continues, Pierre moves to a run-down, dirty part of London, a city he is told is called `London', and he comes to believe that this new town he lives in is ugly, not realizing that ‘London’ refers to the same city as ‘Londres’. Does Pierre now \textit{believe that London is ugly}? Does he also still \textit{believe that London is pretty}? His original representation hasn't changed, but we now want to retract the original ascription. Theorists twist and turn to accommodate this case, as it pulls on Externalist and Internalist intuitions, and it does so in opposing directions that are not eased by hybrid accounts. We will show later that this sort of case is not only common, but central to accounts of learning and development and is easily accommodated when we view the situation from the outside and talk about belief \textit{tokens} instead of belief \textit{types}.

\subsection{Putnam’s Case of Water on Twin Earth}
\vspace*{-1mm}
Next, consider Hilary \citeauthor{putnam1975meaning}'s \citeyear{putnam1975meaning} Twin Earth thought experiment. Twin Earth is just like Earth in every superficial way, with a twin Florida, twins of Earth’s mountains, animals, plants, people, etc. The only difference is that on Twin Earth the watery stuff that fills the oceans and lakes and nourishes its lifeforms is made of some other molecule, XYZ, instead of H$_2$O. As Putnam points out, it would be awkward to say that someone here on Earth and their twin on Twin Earth share \textit{the concept WATER}, even if their internal mental entities are identical.
When the earthling entertains her concept WATER, she is thinking about H$_2$O; when her twin entertains hers, she is thinking about some other substance, XYZ. Again, our Internalist intuitions pull us to say they have the same concept, but our Externalist ones make them distinct. Hybrid accounts don't ease the pain so much as describe it.

\subsection{Putnam’s Case of Jade} 
\vspace*{-1mm}
In the same paper, \citeA{putnam1975meaning} describes the case of jade, a set of stones with similar superficial appearances, but which in fact divide into two very different underlying mineral structures. Suppose we have someone, Peter, who doesn’t know this. Can we, with our superior knowledge, ascribe to Peter \textit{the belief that jade is green}?  It is awkward to do so, at least without heavy elaboration. Consider Mary, a mineral scientist who is well-versed in the two kinds of stone, representing them with their corresponding scientific terms, ‘jadeite’ and ‘nephrite’. Can we say of Mary that she has \textit{the belief that jade is green}? As for the English word, ‘jade’, what is its meaning? Does it fail to have a meaning since using it presupposes, falsely, that there is a uniform mineral kind that it picks out? Or does it represent a disjunctive category, \textit{jadeite or nephrite}, unbeknownst to most users? Again, our intuitions pull us in multiple directions.

On the flip side, it has come to our attention that rubies and sapphires, while different in color, are in fact the same mineral, known as \textit{corundum}. It is only because of impurities, traces of chromium versus iron and titanium, that they reflect light to give their reddish or bluish appearance \cite{ward2003rubies}. Knowing this, can we ascribe to each other \textit{the belief that sapphires are blue}?



\section{Beliefs and Concepts with VW CogSci}

As Timothy \citeA{williamson2007philosophy} argued in his influential book, a major problem with thought experiments is that they are coarse descriptions with a lot of details left to be filled out by our own imaginations, often produced by our prior theories. In Kripke’s case of Pierre, we assume initially that he has only one representation for the city of London. Later in the scenario, this no longer holds, and our belief ascriptions change. One key advantage of VW CogSci for thought experiments is that it forces a more complete and explicit fleshing-out of the scenarios under consideration. Next, we will see how the method can help resolve philosophical puzzles like the ones described in the previous section. Moreover, we will see how the method allows us to track the development of mental representations as agents gain experience with entities in an environment and with other agents who share that environment, much in the spirit of the Constructivist approach described by \citeA{gopnik-wellman-constructivism}.

\begin{figure}[h!]
    \centering
\includegraphics[width=.5\textwidth]{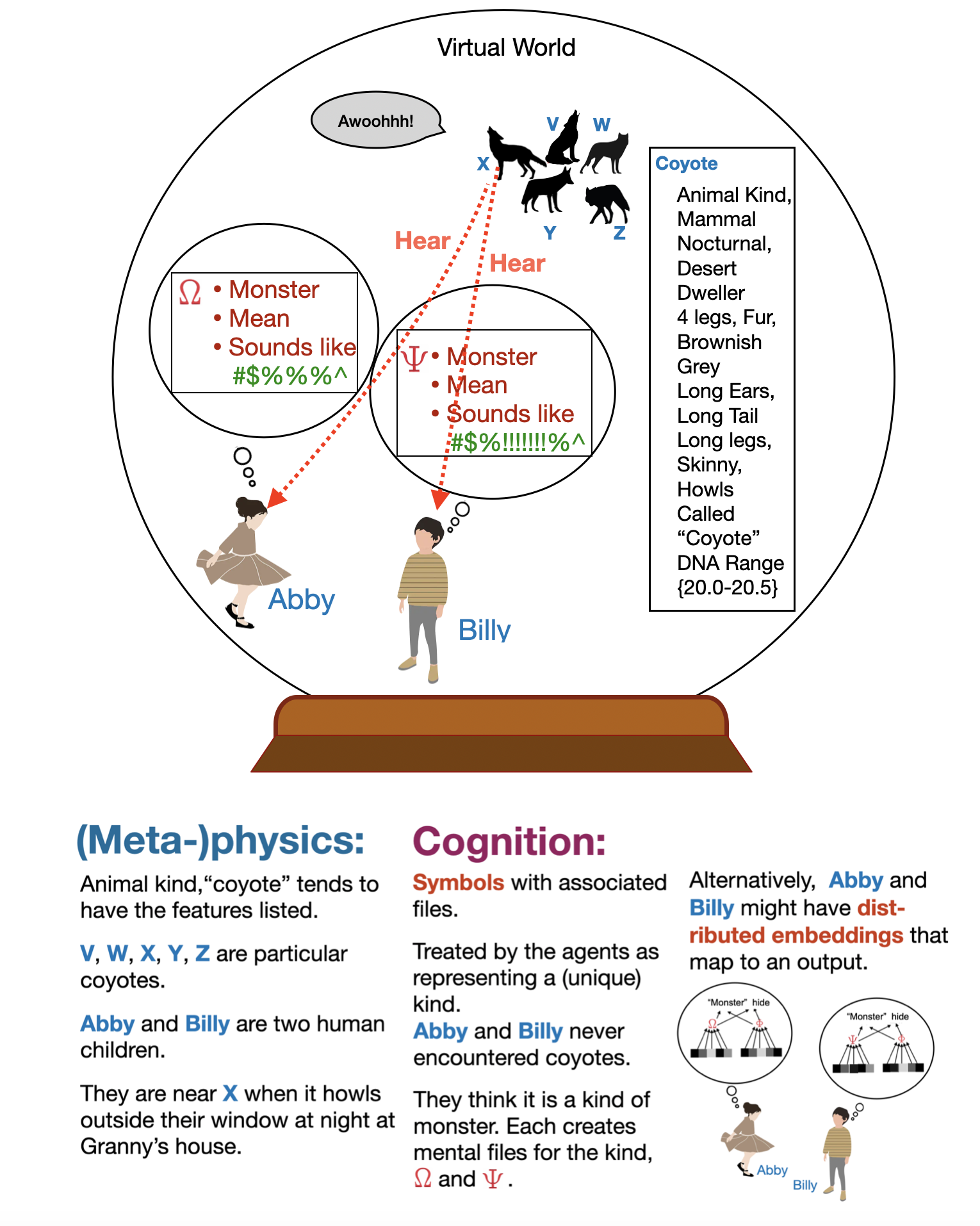}
    \vspace*{-4mm}
    \caption{A virtual world in which Abby and Billy hear coyotes, and each form a concept token for the kind.}
    \label{fig:abby-billy}
    \vspace{-2mm}
\end{figure}

Let’s suppose we build a virtual world that we furnish the with instances of various animal kinds --- dog, wolf, coyote, cat, tiger, such that, e.g., any entity that is a coyote is likely to be furry, tends to howl, has lungs, and so on (see Figure~\ref{fig:abby-billy}).  Such essentialism, with objective facts about whether an entity is a coyote, may or may not be in the metaphysics of our world, but the point is that we can experiment with this too. 

Next, suppose we build two virtual human toddlers, Abby and Billy,
and in the simulation they hear coyotes outside Granny’s house at night. The coyotes produce virtual sound waves that cause auditory sensations in the children. Suppose they both infer that they are hearing some new kind of entity, and they each create a mental file for this newly detected kind, label the kind $\Omega$ and $\Psi$ respectively, and store what they believe to be the likely features of the kind.\footnote{This variation of the labeled-files view of concepts follows the Baptism model proposed by \citeA{Oved2015-OVEHFA}.} (Alternatively, we could build Abby and Billy with a Neural Network architecture, in which case they might form distributed embeddings instead of symbols and files, as shown in the figure.)
    
Already at this point, theorists might begin to argue over whether Abby and Billy have \textit{the COYOTE concept}, or \textit{the belief that coyotes are monsters}, but it’s not clear what that adds to our understanding. VW CogSci gives us a full view from the outside, so we can simply describe Abby’s and Billy’s token labeled files, $\Omega$ and $\Psi$ (or their token distributed embeddings), run the simulation, and observe their attempts to coordinate with other agents and the entities in their world. Abby and Billy have very little information about coyotes, and some of it is false, yet they have representations that allow them to think about what in fact are coyotes, and add knowledge about coyotes as they encounter more of them.

    \begin{figure}[h!]
    \centering
\includegraphics[width=.49\textwidth]{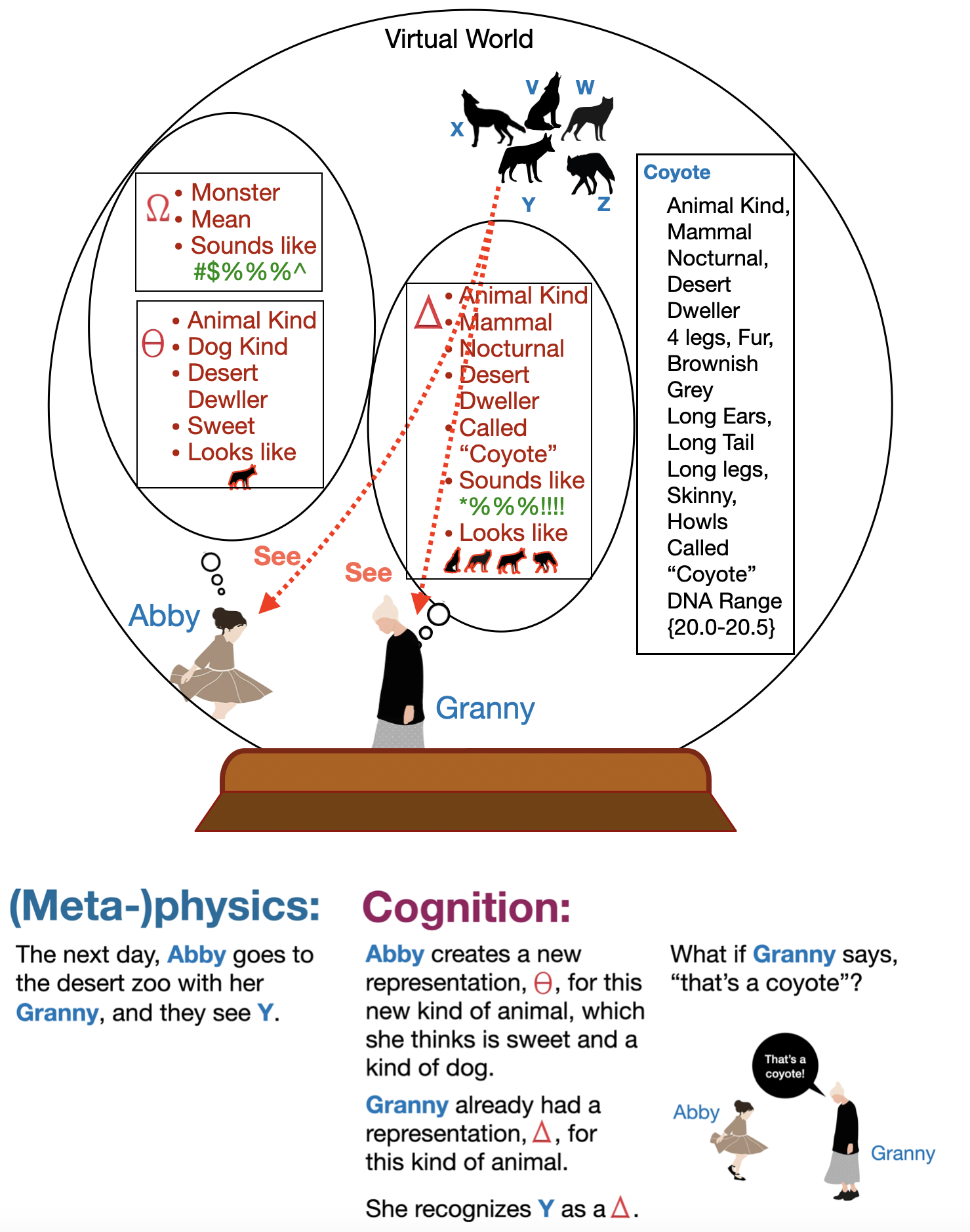}
    \vspace{-4mm}
    \caption{Granny takes Abby to the desert zoo and they see coyotes. Abby creates a new file and label for what she thinks is a new kind of entity.}
    \label{fig:granny-abby}
    \vspace*{-2mm}
    \end{figure}

Next, suppose that Granny takes Abby to the desert zoo the following day, and they see some skinny-legged, furry, dog-like animals which in fact are coyotes. Not knowing these are the same kind of entity as what she heard the night before, Abby creates a new file, $\theta$, (or distributed network) for this category (see Figure~\ref{fig:granny-abby}).  Notice that Abby is a lot like Pierre in \citeauthor{kripke1979puzzle}'s \citeyear{kripke1979puzzle} example; she has two representations that, unbeknownst to her, refer to the same thing. Granny already had encountered coyotes many times before so she already had a labeled file for coyotes, $\Delta$, and recognized the instances.

Do Abby and Granny share \textit{the concept COYOTE}? Does Abby have \textit{the belief that coyotes are mean monsters} or \textit{the belief that coyotes are sweet dogs}? Again, the answers aren’t obvious. Suppose Granny tells Abby that these animals are called ‘coyotes’? Would she then know \textit{what coyotes are}? Would she know \textit{the meaning of the word ‘coyote’}? Abby would presumably still have two mental files which she treats as representing two different kinds of entity. We don’t have English words for her two categories, so we aren’t able to distinguish them with ordinary language. Again, our stance is that it isn’t helpful to try to settle such matters. With the whole picture from the outside, we can simply consider their respective concept and belief \textit{tokens} and observe how easily they coordinate with each other and the entities in their world. 

\begin{figure}[h!]
    \centering
\includegraphics[width=.5\textwidth]{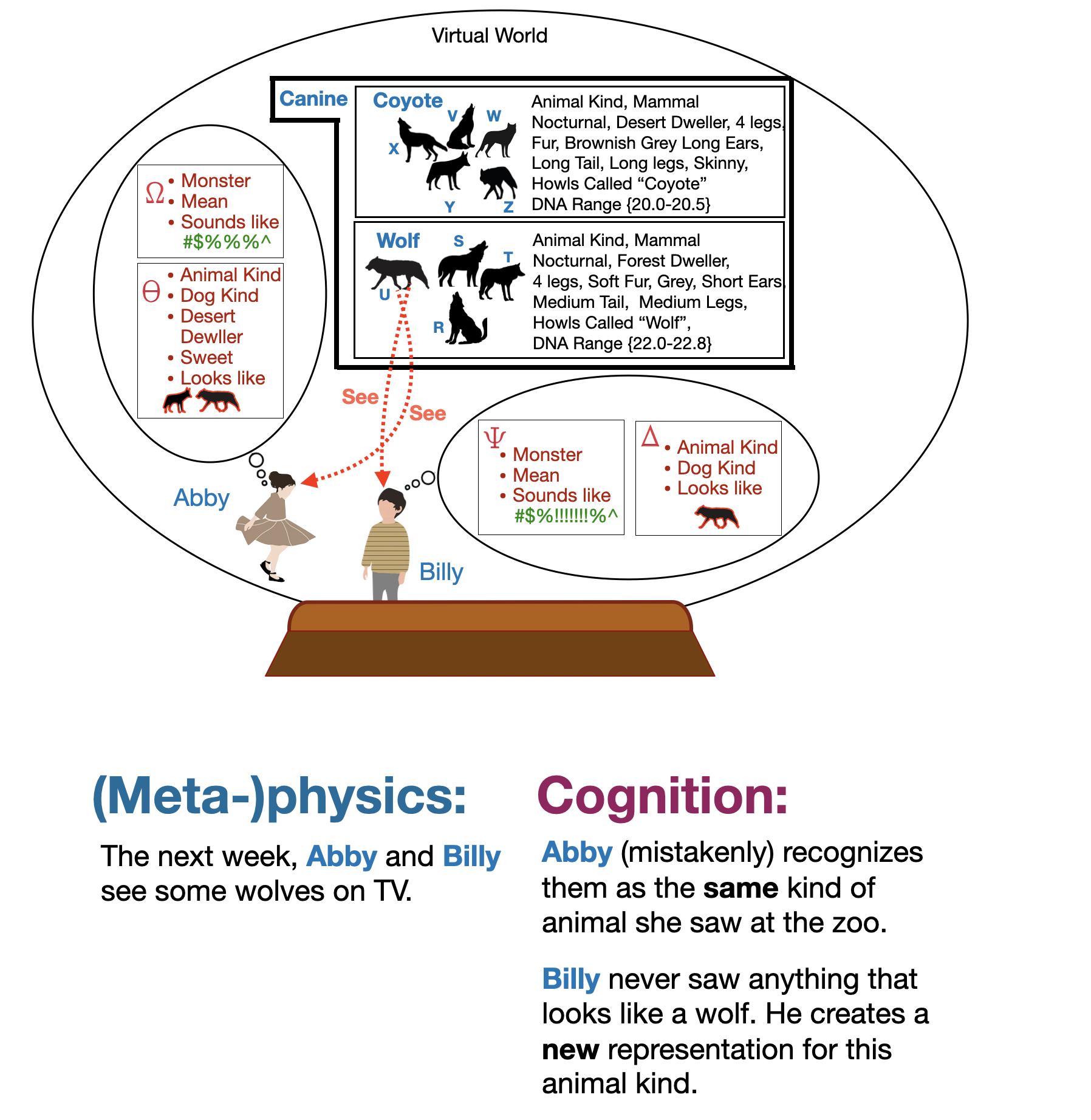}
    \vspace{-4mm}
    \caption{Abby and Billy see wolves on TV. Abby adds details to one of her files while Billy creates a new one.}
    \label{fig:abby-billy-tv}
\end{figure}

We can make matters worse by supposing that Abby and Billy later see what in fact are wolves on TV (Figure~\ref{fig:abby-billy-tv}). Suppose Abby thinks they look like the coyotes she saw at the zoo, so she adds them to her $\theta$ file.  Billy has never encountred this appearance, so he creates a  new file, labeling it $\Delta$. Now Abby’s $\theta$ is a lot like JADE in Putnam’s (1975) example; she treats as one animal-kind what in fact are two kinds.  

Do either Abby or Billy have \textit{the belief that wolves are furry}? Do they share \textit{the concept WOLF}? Is there any sense in which Billy and Granny \textit{share} a concept, given they both have files they label $\Delta$? Again, with the full picture from the outside, it’s not helpful to try to answer these questions. Abby and Billy are children simply trying to learn about their world, carving its joints as best they can as they go along. At this stage in their learning, their carvings fail to correspond to the objective joints in their world, but as we run the simulation and they continue to gain more experience, we will be able to observe whether their models become more aligned with their world. Hopefully Granny will also continue to learn as she ages. She might discover that what people call ‘coyotes’ in her world in fact divide into two different species that cannot mate and have deep biological distinctions. We need room for such development in our theory of beliefs and concepts. These mistakes and revisions are expected, even healthy, and we can fully describe such cognitive development by appeal to the agents' respective concept and belief tokens.

\begin{figure}[h!]
    \centering
    \includegraphics[width=.37 \textwidth]{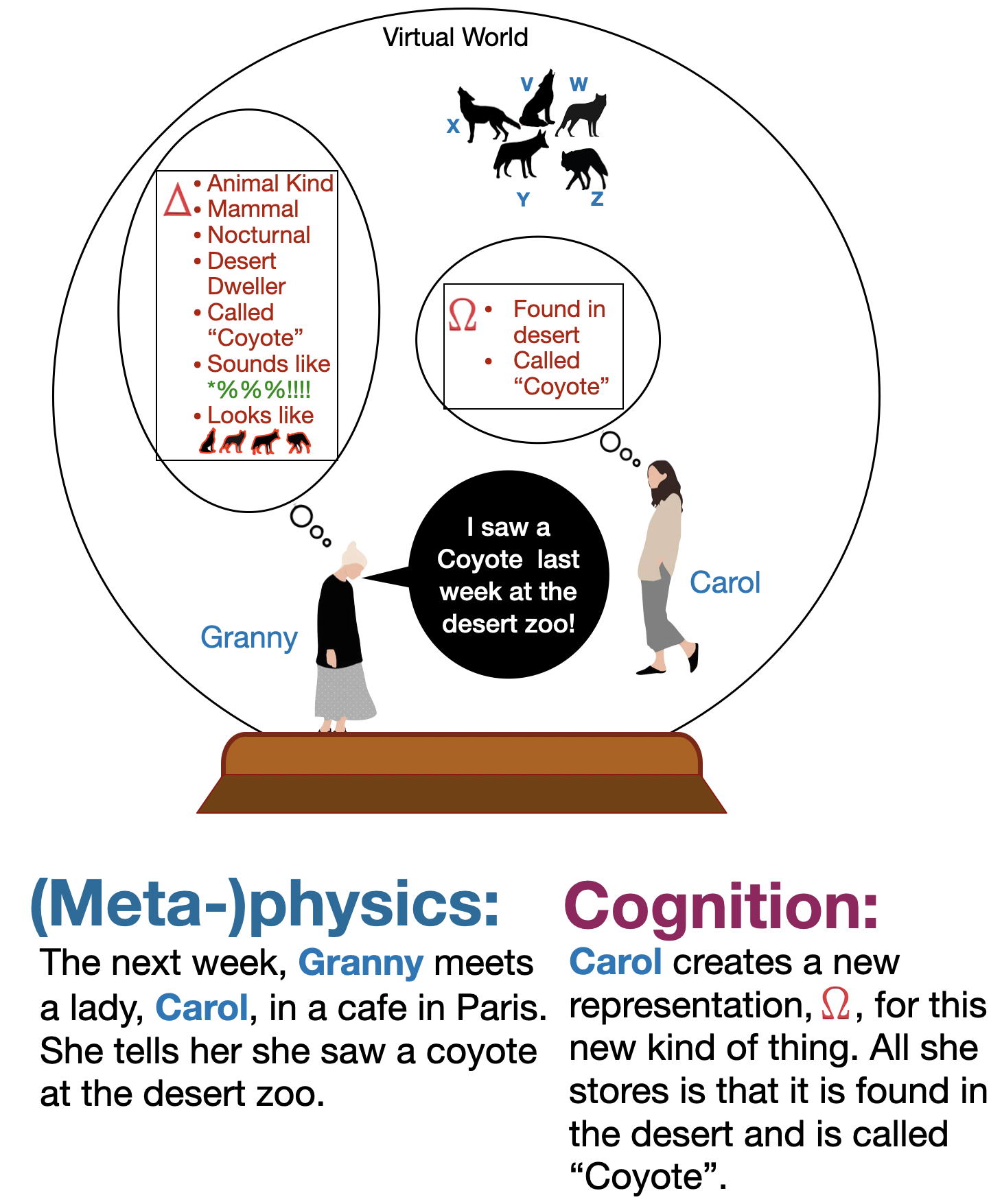}
    \vspace*{-2mm}
    \caption{Granny says to Carol, “I saw a coyote at the desert zoo”. Carol creates a file for the kind.}
    \label{fig:granny-carol}
\end{figure}

Finally, suppose that Granny goes to Paris the next week and meets a woman named Carol. Excited to share about her life in the desert, Granny says to Carol, “I saw a coyote last week at the desert zoo” (see Figure~\ref{fig:granny-carol}). Suppose that Carol infers that ‘coyote’ is the name for some kind of thing that is found in the desert. She creates a file, labels it $\Omega$, and stores the minimal information she has. 
\footnote{This situation is similar to the one described by \citeA{putnam1975meaning} for the English terms ‘Elm’ and ‘Beech’, for which most people have very few associated representations, besides the belief that they name two different kinds of tree.} 
Does Carol have \textit{the COYOTE concept}? She is able to think about what in fact are coyotes, and she can now learn more about them, by asking, e.g., “Are coyotes animals?”, or pointing at something and asking “Is that a coyote?”. But as for \textit{the COYOTE concept}, there is no added value in deciding whether she has it. Notice that if we replace the word
‘coyote’ in this example with an \textit{empty name}, like  ‘witch’, ‘ghost’, or ‘vampire’, we would have no trouble describing the representations as tokens. 

The scenarios described above are typical of the learning process for young children as they encounter new entities in their world. VW CogSci allows us to step outside both the mind and the world in question, so we can eliminate problematic talk about belief and concept \textit{types}, while keeping their tokens. We can then run simulations of Abby and Billy interacting with the entities in their world, and observe how their representations shift, merge, split, how well they correspond to their worlds, and how easily the agents interact with one another through language and gesture. In hindsight, perhaps the real puzzle is why philosophers have been twisting themselves into pretzels for millennia to describe mismatches between an agent's carving of a world and an objective one.

    \vspace*{-2mm}
\section{Conclusions}

This paper gave philosophical motivations for a method we call \textit{Virtual World Cognitive Science} (VW CogSci), in which researchers use virtual embodied agents that are embedded in virtual worlds to explore questions in the field of Cognitive Science. It then showed how the method can be used to dissolve many philosophical puzzles about mental and linguistic representation and test complex theories about the development of such representations. \newline 

After describing the method of VW CogSci, we defended it on the basis of (1) the virtues of computational modeling in the articulation and testing of complex theories in science; (2) the view that mental and linguistic representations are best understood in part by appeal to an agent's sensori-motor interactions with its (assumed) environment; and (3) the claim that the science of the mind must go beyond actual human and animal minds in our world, to include accounts of what kinds of minds are possible and in what kinds of worlds. Cognitive Science, just as \textit{any} science, seeks models that posit a set of entities and regularities that explain our observations, support counterfactuals, and can be tested by interventions. A full model of the mental will thus include relationships between variables in minds, bodies, and worlds. 

We then turned to mental and linguistic representations and showed how VW CogSci adds rigor to their study. First, we showed that by taking a
god's-eye view, the method eliminates the need for problematic talk of belief and concept \textit{types}, such as \textit{the belief that cats are silly}, and \textit{the concept CAT}, while preserving the explanatory power of belief and concept \textit{tokens} in individual cognizers’ minds. Second, we showed how the method allows for the study of various possible minds in various possible worlds to explore questions about the nature, meaning, and development of mental representations by playing out their complex interactions in a simulation rather than trying to track them by armchair analysis.

\section{Acknowledgments}

This work was supported in part by the National Science Foundation (NSF) on grant IIS 2033938 to Boston College and grant IIS 2033932 to Brandeis University, and by the U.S. Army Research Office (ARO) on grant W911NF-23-1-0031 to Colorado State University. The views expressed herein do not reflect the official position of the U.S. Government. We'd also like to thank our anonymous reviewers as well as David Barner, Gedeon Deák, Ian Fasel, Gail Heyman, Terry Horgan, Mengguo Jing, Joshua Knobe, Carlos Montemayor, Seth Neiman, Shaun Nichols, and Matthew Stone for helpful discussion and comments on earlier drafts. Any errors or omissions are the responsibilities of the authors.

\bibliographystyle{apacite}

\setlength{\bibleftmargin}{.125in}
\setlength{\bibindent}{-\bibleftmargin}

\bibliography{CogSci_Template}

\end{document}